\documentclass[letterpaper]{article}
\usepackage{uai2019}
\usepackage[margin=1in]{geometry}

\usepackage{microtype}
\usepackage{graphicx}
\usepackage{subfigure}
\usepackage{booktabs} 

\usepackage{hyperref}

\usepackage{amsfonts}

\usepackage{amsmath}
\usepackage{amsthm}
\usepackage{bm}
\usepackage{color}
\usepackage{bbm}
\usepackage{color,graphicx}
\usepackage{natbib}
\usepackage{enumerate}
\usepackage{multirow}
\usepackage{cases}

\usepackage{comment}

\newcommand{\E}{\mathbb{E}}

\usepackage{multicol}
\usepackage{subfigure}

\graphicspath{{../figures/}}

\newcommand{\cost}{\text{cost}}

\newcommand{\VOIE}{\text{VOI}^\emptyset}
\newcommand{\qKGg}{\text{\it d-KG}}
\newcommand{\taKG}{\text{taKG}}
\newcommand{\taKGE}{\text{taKG}^\emptyset}

\newtheorem{lemma}{Lemma}

\newtheorem{proposition}{Proposition}
\newtheorem{corollary}{Corollary}
\newtheorem{theorem}{Theorem}
\numberwithin{equation}{section}

\let\Section\S
\renewcommand{\S}{\mathcal{S}}

\newcommand{\eqn}[1]{(\ref{eqn:#1})}

\newcommand{\sectn}[1]{\Section\ref{#1}}

\newcommand{\sect}[1]{\ref{sect:#1}}

\newcommand{\Y}{\mathcal{Y}}
\newcommand{\loss}{L}

\usepackage{color}
\usepackage[dvipsnames]{xcolor}
\newcommand{\xxcomment}[4]{\textcolor{#1}{[$^{\textsc{#2}}_{\textsc{#3}}$ #4]}}

\newcommand{\agw}[1]{\xxcomment{red}{A}{W}{#1}}

\usepackage{times}

\title{Practical Multi-fidelity Bayesian Optimization for Hyperparameter Tuning}

\author{
Jian Wu \quad
Saul Toscano-Palmerin \quad
Peter I. Frazier \quad
Andrew Gordon Wilson \\
Cornell University
}

\begin{document}

\maketitle

\begin{abstract}

Bayesian optimization is popular for optimizing time-consuming black-box objectives.
Nonetheless, for hyperparameter tuning in deep neural networks,
the time required to evaluate the validation error for even a few hyperparameter settings remains a bottleneck.
Multi-fidelity optimization promises relief using cheaper proxies to such objectives ---  for example, validation error for a network trained using a subset of the training points or fewer iterations than required for convergence.
We propose a highly flexible and practical approach to multi-fidelity Bayesian optimization, 
focused on efficiently optimizing hyperparameters for iteratively trained supervised learning models.
We introduce a new acquisition function, the trace-aware knowledge-gradient, which efficiently leverages both multiple continuous fidelity controls and trace observations --- values of the objective at a sequence of fidelities, available when varying fidelity using training iterations. 
We provide a provably convergent method for optimizing our acquisition function
and show it outperforms state-of-the-art alternatives for  hyperparameter tuning of deep neural networks and large-scale kernel learning. 

\end{abstract}

\section{INTRODUCTION}
In hyperparameter tuning of machine learning models, we seek to find hyperparameters $x$ in some set ${A} \subseteq \mathbb{R}^d$ to minimize the validation error $f(x)$, i.e., to solve
\begin{equation}
\min_{x\in\mathbb{A}} f(x)
\label{eqn:min_f}
\end{equation}
Evaluating $f(x)$ can take substantial time and computational power \citep{bergstra2012random}
and may not provide gradient evaluations. Bayesian optimization, which requires relatively few 
function evaluations, provides a compelling approach to such optimization problems \citep{jones1998efficient, snoek2012practical}.

As the computational expense of training and testing a modern deep neural network for a single set of hyperparameters has grown, researchers have sought to supplant some evaluations of $f(x)$ with computationally inexpensive low-fidelity approximations.  
Conceputally, an algorithm can use low-fidelity evaluations to quickly identify a smaller set of promising hyperparameter settings, and then later focus more expensive high-fidelity evaluations within this set to refine its estimates.

Pioneering multi-fidelity approaches focused on hyperparameter tuning for deep neural networks
include the Bayesian optimization methods FaBOLAS \citep{klein2016fast,klein2015towards}, Freeze-Thaw Bayesian Optimization \citep{swersky2014freeze}, BOCA \citep{kandasamy2017multi}, predictive entropy search for a single continuous fidelity \citep{mcleod2017practical}, early-stopping SMAC \citep{domhan2015speeding}, and
Hyperband \citep{li2016hyperband}.
This work builds on
earlier multi-fidelity optimization approaches \citep{huang2006sequential,lam2015multifidelity,poloczek2016multi} focused on low-fidelity approximation of physics-based computational models.

These validation error approximations 
perform the same training and testing steps as in standard Bayesian optimization, but control fidelity with fewer training iterations than required for convergence, fewer training data points, or fewer validation data points. 
These approximations present unique opportunities not typically considered in the multifidelity literature, even within the portion focused on hyperparameter tuning. 
First, we observe a full \emph{trace} of performance with respect to training iterations, rather than just a single performance value at the chosen fidelity.
Indeed, training with $s$ iterations produces evaluations of the low-fidelity approximation for {\it all} training iterations less than or equal to $s$.  
Second, by caching state after completing $s$ iterations, we can significantly reduce computation time when later evaluating for $s'>s$ evaluations.  This allows quickly evaluating low-fidelity approximations to the validation error for many hyperparameter settings, then later returning to those most promising hyperparameter settings to cheaply obtain more accurate observations.
Third, we may simultaneously alter fidelity along several continuous dimensions (iterations, training data, validation data), rather than modifying one continuous fidelity control or choosing from among a discrete set of ambiguously related fidelities.


In this paper, we propose the trace-aware knowledge gradient (taKG) for Bayesian optimization with multiple fidelities. taKG is distinctive in that it leverages both trace information and multiple fidelity controls at once, efficiently selecting training size, validation size, number of training iterations, and hyperparameters to optimize. Moreover, we provide a provably-convergent method for maximizing this acquisition function.
taKG addresses the challenges presented by trace observations by considering the reduced cost of adding iterations at a previously evaluated point, and using an intelligent selection scheme to choose a subset of the observed training iterations to include in inference.
Additionally, taKG can be used in either a batch or sequential setting, and can also efficiently leverage gradient information if it is available.

We present two variants of our trace-aware knowledge-gradient acquisition function, one for when the cost of sampling is substantial over the whole fidelity space (even when using few training points or iterations), and the other for when the cost and value of information vanish as fidelity decreases to 0.  The first form we refer to simply as taKG, and the second as 0-avoiding taKG ($\taKGE$) because it avoids the tendency of multi-fidelity other methods to measure repeatedly at near-0 fidelities even when these low fidelities provide almost no useful information.  Alternative approaches \citep{mcleod2017practical,klein2016fast} add and tune a fixed cost per sample to avoid this issue, while $\taKGE$ does not require tuning. 

Furthermore, we present a novel efficient method to optimize these acquisition functions, even though they cannot be evaluated in closed form.
 This method first constructs a stochastic gradient estimator which it then uses within multistart stochastic gradient ascent. We show that our stochastic gradient estimator is unbiased and thus asymptotically consistent, and the resulting stochastic gradient ascent procedure converges to a local stationary point of the acquisition function.
 
Our numerical experiments demonstrate a significant improvement over state-of-the-art alternatives, including FaBOLAS \citep{klein2016fast,klein2015towards}, Hyperband \citep{li2016hyperband}, and 
BOCA \citep{kandasamy2017multi}. Our approach is also applicable to problems that do not have trace observations, but use continuous fidelity controls, and we additionally show strong performance in this setting.

In general, efficient and flexible multifidelity optimization is of crucial practical importance, as evidenced by growing momentum in this research area. Although Bayesian optimization has shown great promise for tuning hyperparameters of machine learning algorithms, computational bottlenecks have remained a major deterrent to mainstream adoption. With taKG, we leverage crucial trace information, while simultaneously providing support for several fidelity controls, providing remarkably efficient optimization of expensive objectives. This work is intended as a step towards the renewed practical adoption of Bayesian optimization for machine learning.

\section{THE taKG AND $\taKGE$ ACQUISTION FUNCTIONS}
\label{sect:taKG}
In this section we define the trace-aware knowledge-gradient acquisition function.
\sectn{sec:obj} defines our formulation of multi-fidelity optimization with traces and continuous fidelities, along with our inference procedure.
\sectn{sect:valuing} describes a measure of expected solution quality possible after observing a collection of fidelities within a trace.
\sectn{sect:voi} uses this measure to define the $\taKG$ acquisition function,
and \sectn{sect:mvoi} defines an improved version, $\taKGE$, appropriate for settings in which the 
the cost and value of information vanish together (for example, as the number of training iterations declines to 0).
\sectn{sect:optimize} then presents a computational approach for maximizing the $\taKG$ and $\taKGE$ acquisition functions and theoretical results justifying its use.  \sectn{sect:warm-start} discusses warm-starting previously stopped traces, and \sectn{sect:extensions} briefly discusses generalizations to batch and derivative observations.




\subsection{Problem Setting}
\label{sec:obj}
We model our objective function and its inexpensive approximations by a real-valued function $g(x, s)$ where our objective is $f(x) := g(x, 1)$ and $s \in [0,1]^m$ denotes the $m$ fidelity-control parameters.   (Here, $1$ in $g(x,1)$ is a vector of $1$s.)  
We assume that our fidelity controls have been re-scaled so that $1$ is the highest fidelity and $0$ the lowest.  
$g(x,s)$ can be evaluated, optionally with noise, at a cost depending on $x$ and $s$. 

We let $B(s)$ be the additional fidelities observed for free when observing fidelity $s$.  Although our framework can be easily generalized, we assume that $B(s)$ is a cross product of sets of the form either $[0,s_i]$ (\emph{trace fidelities}) or $\{s_i\}$ (\emph{non-trace fidelities}). 
We let $m_1$ denote the number of trace fidelities and $m_2$ the number of non-trace fidelities.
We also assume that the cost of evaluation is non-decreasing in each component of the fidelity.

For example, consider hyperparameter tuning with $m=2$ fidelities: first is the number of training iterations; 
second is the amount of training data.  
Each is bounded between $0$ and some maximum value, and $s_i \in [0,1]$ specifies training iterations or number of training data points as a fraction of this maximum value.
Then, $B(s) = [0,s_1] \times \{s_2\}$, because we observe results for the number of training iterations ranging from $0$ up to the number evaluated.  
If the amount of validation data is another trace fidelity, we would have: $B(s) = [0,s_1] \times \{s_2\} \times [0,s_3]$. 

We model $g$ using Gaussian process regression jointly over $x$ and $s$, assuming that observations are perturbed by independent normally distributed noise with mean $0$ and variance $\sigma^2$.  Each evaluation consists of $x$, a vector of fidelities $s$, and a noisy observation of $g(x,s')$ for each fidelity $s'$ in $B(s)$.  For computational tractability, in our inference, we will choose to retain and incorporate observations only from a subset $\mathcal{S} \subseteq B(s)$ of these fidelities with each observation.  After $n$ such evaluations, we will have a posterior distribution on $g$ that will also be a Gaussian process, and whose mean and kernel we refer to by $\mu_n$ and $K_n$.  We describe this inference framework in more detail in the supplement.

We model the logarithm of the cost of evaluating $g$ using a separate Gaussian process, updated after each evaluation,
and let $\cost_n(x,s)$ be the predicted cost after $n$ evaluations.
We assume for now that the cost of evaluation does not depend on previous evaluations, and then discuss later in \sectn{sect:warm-start} an extension to warm-starting evaluation at higher fidelities using past lower-fidelity evaluations.

\subsection{Valuing Trace Observations}
\label{sect:valuing}

Before defining the $\taKG$ and $\taKGE$ acquisition functions, we define a function $\loss_n$ that quantifies the extent to which observing trace information improves the quality of our solution to \eqref{eqn:min_f}.

Let $\E_n$ indicate the expectation with respect to the posterior $\mathbb{P}_n$ after $n$ evaluations.  Given any $x$ and set of fidelities $\S$, 
we will define a function $\loss_n(x,\S)$ to be the expected loss (with respect to the time-$n$ posterior) of our final solution to \eqref{eqn:min_f} if we are allowed to first observe $x$ at all fidelities in $\S$.

To define this more formally, let $\Y(x,\S)$ be a random vector comprised of observations of $g(x,s)$ for all $s \in \S$.  Then, the conditional expected loss from choosing a solution $x'$ to \eqref{eqn:min_f} after this observation is $\E_n\left[g(x',1) \mid \Y(x,\S)\right]$.  
This quantity is a function of $x$, $\S$, $\Y(x,\S)$, and the first $n$ evaluations, and 
can be computed explicitly using formulas from GP regression given in the supplement.

We would choose the solution for which this is minimized, giving a conditional expected loss of $\min_{x'} \E_n\left[g(x',1) \mid \Y(x,\S) \right]$. This is a random variable under the time-$n$ posterior whose value depends on $\Y(x,\S)$.
We finally take the expected value under the time-$n$ posterior to obtain $\loss_n(x,\S)$:
\begin{equation} 
\begin{split}
&\loss_n(x,\S) := \E_n\left[ \min_{x'} \E_n\left[g(x',1) \mid \Y(x,\S) \right]\right] \\
&\!=\!\int\!\mathbb{P}_n\!\left(\Y(x,\S)\!=\!y\right) \min_{x'} \E_n\left[g(\!x'\!,\!1\!)\!\mid\!\Y(\!x\!,\!\S)\!=\!y \right]\, dy,
\end{split}
\end{equation}
where the integral is over all $y \in \mathbb{R}^{|\S|}$.

We compute this quantity using simulation.  To create one replication of this simulation we first simulate $(g(x,s) : s \in S)$ from the time-$n$ posterior.  We then add simulated noise to this quantity to obtain a simulated value of $\Y(x,\S)$.  We then update our posterior distribution on $g$ using this simulated data, allowing us to compute $\E_n\left[g(x',1) \mid \Y(x,\S)\right]$ for any given $x'$ as a predicted value from GP regression.
We then use continuous optimization method designed for inexpensive evaluations with gradients (e.g., multi-start L-BFGS) to optimize this value, giving one replication of $\min_{x'} \E_n\left[g(x',1) \mid \Y(x,\S)\right]$.  We then average many replications to give an unbiased and asymptotically consistent estimate of $\loss_n(x,\S)$.

In a slight abuse of notation, we also define $\loss_n(\emptyset) = \min_{x'} \E_n\left[g(x',1) \right]$.  This is the minimum expected loss we could achieve by selecting a solution without observing any additional information.
This is equal to $\loss_n(x,\emptyset)$ for any $x$.

The need to compute $\loss_n(x,\S)$ via simulation will present a challenge when optimizing acquisition functions defined in terms of it.  Below, in \sectn{sect:optimize} we will overcome this challenge via a novel method for simulating unbiased estimators of the gradient of $\loss_n(x,\S)$ with respect to $x$ and the components of $\S$.  First, however, we define the $\taKG$ and $\taKGE$ acqisition functions.

\subsection{Trace-aware Knowledge Gradient (taKG)}
\label{sect:voi}


The $\taKG$ acquisition function will value observations of a point and a collection of fidelities according to the ratio of the reduction in expected loss (as measured using $\loss_n$) that it induces, to its computational cost.

While evaluating $x$ at a fidelity $s$ in principle provides observations of $g(x,s')$ at all $s' \in  B(s)$, we choose to retain and include in our inference only a subset of the observed fidelities $\S \subseteq B(s)$.  This reduces computational overhead in GP regression.  In our numerical experiments, we take the cardinality of $\S$ to be either 2 or 3, though the approach also allows increased cardinality. 

With this in mind, the $\taKG$ acquisition function at a point $x$ and set of fidelities $\S$ at time $n$ is
\begin{equation*}
\taKG_n(x, \mathcal{S}) := \frac{\loss_n(\emptyset) - \loss_n(x,\S)}{\cost_n(x, \max \S)},
\end{equation*}
where we also refer to the numerator as the {\it value of information} \citep{Ho66},
$\text{VOI}_n(x,\S) := \loss_n(\emptyset) - \loss_n(x,\S)$.
Thus, $\taKG$ quantifies the value of information per unit cost of sampling.

The cost of observing at all fidelities in $\S$ is taken here to be the cost of evaluating $g$ at a vector of fidelities equal to the elementwise maximum, $\max \mathcal{S} := (\max_{s \in \mathcal{S}} s_i: 1 \le i \le m)$.  
This is the least expensive fidelity at which we could observe $\S$. 


The taKG algorithm chooses to sample at the point $x$, fidelity $s$, and additional lower-fidelity point(s) $\S \setminus \{s\}$ to retain that jointly maximize the $\taKG$ acquisition function, among all fidelity sets $\S$ with limited cardinality $\ell$.

\begin{equation}
\max_{x,s,\mathcal{S}:\mathcal{S}\subseteq B\left(s\right),\left|\mathcal{S}\right|=\ell,s\in\mathcal{S}} \taKG_n \left(x, \mathcal{S} \right).
\label{eqn:max-taKG}
\end{equation}
This is a continuous optimization problem whose decision variable is described by $d + \ell m_1 + m_2$ real numbers. $d$ describe $x$, $m=m_1+m_2$ describe $s$, and $(\ell-1)m_1$ describe $\S \setminus \{s\}$.

\subsection{0-avoiding Trace-aware Knowledge Gradient ($\taKGE$)}
\label{sect:mvoi}
The $\taKG$ acquisition function uses the value of information per unit cost of sampling.  
When the value of information and cost become small simultaneously, as when we shrink training iterations or training data to 0 in hyperparameter tuning, this ratio becomes sensitive to misspecification of the GP model on $g$. We first discuss this issue, and then develop a version of $\taKG$ for these settings.

To understand this issue, we first observe the value of information for sampling $g(x,s)$, for any $s$, is strictly positive when the kernel has strictly positive entries.
\begin{proposition}
If the kernel function $K_n((x, s), (x', 1)) > 0$ for any $x, x' \in \mathbb{A}$, then for any $x \in \mathbb{A}$ and any $s\in[0,1]^m$, $\text{VOI}_n(x, \left\{ s\right\}) > 0$.
\label{prop:positive}
\end{proposition}

Proposition~\ref{prop:positive} holds even if $s=0$, or has some components set to 0.
Thus, if the estimated cost at such extremely low fidelities is small relative to the (strictly positive) value of information there, $\taKG$ may be drawn to sample them, even though the value of information is small.  We may even spend a substantial portion of our budget evaluating $g(x,0)$ at different $x$.
This is usually undesirable.

For example, in hyperparameter tuning with training iterations as our fidelity, 
fidelity $0$ corresponds to training a machine learning model with no training iterations.
This would return the validation error on initial model parameter estimates.
While this likely provides {\it some} information about the validation error of a fully trained model, 
specifying a kernel over $g$ that productively uses this information from a large number of hyperparameter sets $x$ 
would be challenging.

This issue becomes even more substantial when considering training iterations
and training data together, as we do here, because cost nearly vanishes
as either fidelity vanishes.  Thus, there are many
fidelities at each $x$ that we may be drawn to oversample.

This issue is not specific to \taKG. 
It also occurs in previous literature 
\citep{klein2016fast,mcleod2017practical,klein-bayesopt17}
when using the ratio of information gain to cost in 
an entropy search or predictive entropy search method based on the same predictive model.


To deal with this issue, \citet{klein2016fast,mcleod2017practical} artificially inflate the cost of evaluating at fidelity $0$ to penalize low fidelity evaluations.  Similarly, \citet{klein-bayesopt17} recommends adding a fixed cost to all evaluations motivated by the overhead of optimizing the acquisition function, but then recommends setting this to the same order of magnitude as a full-fidelity evaluation even though the overhead associated with optimizing a BO acquisition function using well-written code and efficient methodology will usually be substantially smaller. As a result, any fixed cost must be tuned to the application setting to avoid oversampling at excelssively small fidelities while still allowing sampling at moderate fidelities.

Here, we propose an alternate solution that we find works well without tuning,
focusing on the setting where the cost of evaluation becomes small as the smallest component in $s$ approaches 0.


We first define $C(s) = \cup_{i=1}^m \{ s' : s'_i = 0, s'_j = s_i\ \forall j\ne i \}$
to be the set of fidelities obtained by replacing one component of $s$ by $0$.
We then let $C(\S) = \cup_{s \in \S} C(s)$ be the union of these fidelities over $s \in \S$.
For example, suppose $s_1$ is a trace fidelity (say, training iterations), $s_2$ is a non-trace fidelity (say, training data size), and $\S = \{ (1/2, 1), (1,1) \}$, corresponding to an evaluation of $g$ at $s=(1,1)$ and retention of the point $(1/2,1)$ from the trace $B((1,1))$.
Then $C(\mathcal{S}) = \{ (0,1), (1/2, 0), (1,0) \}$.  


Fidelities in $C(\S)$ (for any $\S$) are extremely inexpensive to evaluate and provide extremely small but strictly positive value of information.
These, and ones close to them, are ones we wish to avoid sampling, even when $\taKG$ is large.

To accomplish this, we modify our value of information 
$\text{VOI}_n(x,\S) = \loss_n(\emptyset) - \loss_n(x,\S)$
to suppose free observations $\Y(x,s')$ will be provided of these problematic low-fidelity $s'$.
  Our modified value of information will suppose these free observations will be provided to both the benchmark, previously set to $\loss_n(\emptyset)$, and to the reduced expected loss, previously set to $\loss_n(x,\S)$, achieved through observing $x$ at fidelities $\S$.
The resulting modified value of information is
\begin{equation*}
\VOIE_n(x,\S) = \loss_n(x,C(\S)) - \loss_n(x,\S \cup C(\S)) \notag \\
\end{equation*}
We emphasize our algorithm will not evaluate $g$ at fidelities in $C(\S)$.  Instead, it will simulate these evaluations according to the algorithm in \sectn{sect:valuing}.

We define the $\taKGE$ acquisition function using this modified value of information as
\begin{equation}
\taKGE_n(x, \mathcal{S}) = \frac{\VOIE_n(x,\S)}{\text{cost}_n(x , \max \S)}
\label{eqn:takg0}
\end{equation}
To find the point $x$ and fidelity $s$ to sample, we optimize $\taKGE$ over 
$x$, fidelity $s$, and additional lower-fidelity point(s) $\S \setminus \{s\}$ as we did in \eqref{eqn:max-taKG}.

We refer to this VOI and acquisition function as ``0-avoiding,'' because they place 0 value on fidelities with any component equal to 0. 
This prevents sampling at these points as long as the cost of sampling is strictly positive.

Indeed, suppose $s=\max(\S)$ has a component equal to $0$.  Then 
each element in $\S$ will have one component equal to $0$, and $\S \subseteq C(\S)$.
Then $\VOIE_n(x,\S) = \loss_n(x,C(\S)) - \loss_n(x,C(\S)\cup \S) = 0$.
Moreover, the following proposition shows that if $s=\max(\S)$ has all components strictly positive and additional regularity conditions hold, then $\VOIE_n(x,\S)$ is also strictly positive.

\begin{proposition}
If $s=\max(\S)$ has all components strictly positive, our kernel $K_n$ is positive definite, and the hypothesis of Proposition 1 is satisfied for $K_n$ given $g(x,C(\S))$, then $\VOIE_n(x,\S)$ is strictly positive.
\end{proposition}

Thus, maximizing $\taKGE$ will never choose to sample at a fidelity $s$ with a $0$ component.
Additionally, under other regularity conditions (see Corollary~1 in the supplement), $\VOIE_n(x,\S)$ is continuous in $\S$, and so the property that $\VOIE_n(x,\S)=0$ when a component of $s=\max(S)$ is 0 also discourages sampling at $s$ whose smallest component is {\it close} to 0.

\subsection{Efficiently maximizing $\taKG$ and $\taKGE$}
\label{sect:optimize}


The $\taKG$ and $\taKGE$ acquisition functions are defined in terms of a hard-to-calculate function $\loss_n(x,\S)$. 
Here, we describe how to efficiently maximize these acquisition functions using stochastic gradient ascent with multiple restarts.  The heart of this method is a simulation-based procedure for simulating a stochastic gradient of $\loss_n(x,\S)$, i.e., a random variable whose expectation is the gradient of $\loss_n(x,\S)$ with respect to $x$ and the elements of $\S$.

To construct this procedure, we first provide a more explicit expression for $\loss_n(x,\S)$.
Because $\loss_n(x,\S)$ is the expectation of the minimum over $x'$ of $\E_n\left[g(x',1) \mid \Y(x,\S)\right]$, 
we begin with the distribution of this conditional expectation for a fixed $x'$ under the time-$n$ posterior distribution.

This conditional expectation can be calculated with GP regression from previous observations, the new point $x$ and fidelities $\S$, and the observations $\Y(x,\S)$.  This conditional expectation is linear in $\Y(x,\S)$.

Moreover, $\Y(x,\S)$ is the sum of $g(x,\S)$ (which is multivariate normal under the posterior) and optional observational noise (which is independent and normally distributed), and so is itself multivariate normal.
As a multivariate normal random variable, it can be written as the sum of its mean vector and the product of the Cholesky decomposition of its covariance matrix with an independent standard normal random vector, call it $W$.  (The coefficients of this mean vector and covariance matrix may depend on $x$, $\S$, and previously observed data.)  The dimension of $W$ is the number of components in the observation, $|\S|$.

Thus, the conditional expected value of the objective $\E_n\left[g(x',1) \mid \Y(x,\S)\right]$ is a linear function (through GP regression) of another linear function (through the distribution of the observation) of $W$.

We also have that the mean of this conditional expectation is
$\E_n[\E_n[g(x’,1) | \Y(x,\S) ] ] =
\E_n[g(x’,1) ] = \mu_n(x’)$ by iterated conditional expectation.

These arguments imply the existence of a function $\tilde{\sigma}_n(x',x,\S)$ such that
$\E_n[g(x',1) | \Y(x,\S)]
= \mu_n + \tilde{\sigma}_n(x',x,\S) W$ simultaneously for all $x'$.
In the supplement, we show 
$\tilde{\sigma}_n(x',x,\S)= K_n\left((x', 1), x_{\S}\right) (D_{n}^T)^{-1}$
where $x_\mathcal{S} = \{(x, s): s \in \mathcal{S}\}$, and $D_{n}$ is the Cholesky factor of the covariance matrix $K_n\left(x_\mathcal{S}, x_\mathcal{S}\right)+\sigma^2 I$.


Thus, 
\begin{equation*}
\loss_{n}\left(x,\S\right)=\mathbb{E}_{n}\left[\mbox{min}_{x'} \mu^{\left(n\right)}\left(x',1\right)+\tilde{\sigma}_{n}\left(x', x, \S\right)W\right].
\end{equation*}


When certain regularity conditions hold,
\begin{align*}
\nabla \loss_n(x,\S)
&= \nabla \mathbb{E}_{n}\left[\min_{x'}\left(\mu_n\left(x',1\right)+\tilde{\sigma}_{n}\left(x', x, \S\right)W\right)\right]\\
&= \mathbb{E}_{n}\left[\nabla \min_{x'}\left(\mu_n\left(x',1\right)+\tilde{\sigma}_{n}\left(x', x, \S\right)W\right)\right]\\
&= \mathbb{E}_{n}\left[\nabla \left(\mu_n\left(x^*,1\right)+\tilde{\sigma}_{n}\left(x^*, x, \S\right)W\right)\right]\\
&= \mathbb{E}_{n}\left[\nabla \tilde{\sigma}_{n}\left(x^*, x, \S\right)W\right],
\end{align*}
where $x^*$ is a global minimum (over $x'\in \mathbb{A}$) of $\mu_n(x',1) + \tilde{\sigma}_n(x',x,\S)W$, and the gradient in the last line is taken holding $x^*$ fixed even though in reality its value depends on $x$.
Here, the interchange of expectation and the gradient is justified using results from infinitessimal perturbation analysis \citep{l1990unified} and ignoring the dependence of $x^*$ on $x$ is justified using the envelope theorem \citep{milgrom2002envelope}.
We formalize this below in Theorem~\ref{stochastic_gradient}.

\begin{theorem}
\label{stochastic_gradient}
Suppose $\mathbb{A}$ is compact, $\mu_0$ is constant, the kernel $K_0$ is continuously differentiable, and $\mathrm{argmin}_{x'\in\mathbb{A}}\ \mu_{n}(x', 1)+\tilde{\sigma}_n\left(x', x, \S\right)W$ contains a single element almost surely. Then, $\nabla \loss_n(x,\S) = \mathbb{E}_n\left[ \nabla \tilde{\sigma}(x^*,x,\S) W \right]$
\end{theorem}

With this result in place, we can obtain an unbiased estimator of $\nabla \loss_n(x,\S)$ by simulating $W$, calculating $x^*$, and then returning $\nabla \tilde{\sigma}_n(x^*,x,\S) W$.
Using this, together with the chain rule and an exact gradient calculation for $\text{cost}_n(x,\max \mathcal{S})$,
we can then compute stochastic gradients for $\taKG$ and $\taKGE$.

We then use this stochastic gradient estimator within stochastic gradient ascent \citep{kushner2003stochastic}
to solve the optimization problem \eqn{max-taKG} (or the equivalent problem for maximizing $\taKGE$).  The following theorem shows that, under the right conditions, a stochastic gradient ascent algorithm converges almost surely to a critical point of $\taKGE$. Its proof is in the supplement. 

\begin{theorem}
Assume the conditions of Theorem \ref{stochastic_gradient}, $\mathbb{A}$ is a compact hyperrectangle, and $\cost_n(\max \mathcal{S})$
is continuously differentiable and bounded below by a strictly positive constant. In
addition, assume that we optimize $\taKGE$ using a stochastic gradient
ascent method with the stochastic gradient from Theorem \ref{stochastic_gradient} whose stepsize
sequence $\left\{ \epsilon_{t}:t=0,1,\ldots\right\} $ 
satisfies $\epsilon_{t}\rightarrow0$, $\epsilon_{t}\geq0$,
$\sum_{t}\epsilon_{t}=\infty$ and $\sum_{t}\epsilon_{t}^{2}<\infty$.
Then the sequence of points $\left\{ x_{t}, \S_{t}\right\} _{t\geq0}$ from stochastic gradient ascent
converges almost surely to a connected set of stationary points of $\taKGE$.
\label{t:convergence}
\end{theorem}

\subsection{Warm-starting from partial runs}
\label{sect:warm-start}
When tuning hyperparameters using training iterations as a fidelity,
we can cache the state of training after $s$ iterations, for a warm start, and then continue training later up to a larger number of iterations $s'$ for less than $s'$ training iterations would cost at a new $x$.

We assume trace fidelities can be ``warm-started'' while non-trace fidelities cannot.  We also assume the incremental cost of evaluting fidelity vector $s'$ warm-starting from $s$ is the difference in the costs of their evaluations from a ``cold start''.  We model the cost of cold-start evaluation as in \sectn{sec:obj} with a Gaussian process on log(cost).  To obtain training data for this model, costs observed from warm-started evaluations are summed with those of the previous evaluations they continue to approximate what the cold-start cost would be.  We set $\cost_n(x,s)$ to be the difference in estimated cold-start costs if a previous evaluation would allow warm starting, and to the estimated cold start cost if not.

While our approach to choosing $x$ and $s$ to evaluate is to optimize $\taKGE$ as before 
our computational approach from \sectn{sect:optimize} required that $\cost_n(x,s)$ be continuously differentiable (in Theorem~\ref{t:convergence}).  This requirement is not met.
To address this, we modify the way we optimize the $\taKGE$ acquisition function.
(The approach we describe also works for $\taKG$.)

First, we maintain a basket of size at most $b$ of previously evaluated point, fidelity pairs, $(x(j),s(j))$.
For each $j\le b$, we optimize $\taKGE_n(x(j),\S)$ letting $\S$ vary over those sets satisfying two conditions: (1) $|\S| = \ell$; 
(2) $s' \ge s(j)$ componentwise for each $s' \in \S$, 
with equality for non-trace fidelity components. 
Over this set, $\cost_n(x(j),\S)$ is continuously differentiable in $\S$ and the method from \sectn{sect:optimize} can be applied.


We also optimize $\taKGE_n(x,\S)$ over all $x$ and $\S$ with $|\S|=\ell$, but using the estimated cold-start cost function and the method from \sectn{sect:optimize}.

Among the solution to these at most $b+1$ optimization problems, we select the $x$ and $\S$ that provide the largest $\taKGE_n(x,\S)$ at optimality, and evaluate $g$ at $x$ and $\max(\S)$.

We then update our basket.
We first add the $x$ and $\max(\S)$ produced by the optimization not constraining $x$.
If the basket size exceeds $b$, we then remove the $x$ and $s$ whose optimization over $\taKGE_n$ produced the smallest value.
In practice, we set $b=10$.

\subsection{Batch and Derivative Evaluations}
\label{sect:extensions}
$\taKG$ and $\taKGE$ generalize naturally 
following \citet{wu2016parallel} and \citet{wu2017bayesian}
to batch settings where we can evaluate multiple point, fidelity pairs simultaneously and derivative-enabled settings where we observe gradients. 

The batch version uses the same acquisition functions $\taKG$ and $\taKGE$ defined above, but optimizes over a {\it set} of values for $s$, each of which has an associated $\S \in B(s)$ of limited cardinality.

In the derivative-enabled setting, we incorporate (optionally noisy) gradient observations into our posterior distribution directly through GP regression.  We also generalize the $\taKG$ and $\taKGE$ acquisition functions to allow inclusion of gradients of the objective in the set of quantities observed $\Y(x,\S)$ in the definition of $L_n(x,\S)$.

\section{NUMERICAL EXPERIMENTS}
\label{sect:numerical}
We compare sequential, batch, and derivative-enabled $\taKGE$ with benchmark algorithms on synthetic optimization problems (Sect.~\sect{synthetic}), hyperparameter optimization of neural networks (Sect.~\sect{hyperexp}), and hyperparameter optimization for large-scale kernel learning (Sect.~\sect{kernel-learning}).  The synthetic and neural network benchmarks use fidelities with trace observations, while the large-scale kernel learning benchmark does not.  We integrate over GP hyperparameters by sampling $10$ sets of values using the \texttt{emcee} package \citep{foreman2013emcee}.

\subsection{Optimizing synthetic functions} \label{sect:synthetic}
Here, we compare $\taKGE$ against both the sequential and batch versions of the single-fidelity algorithms KG \citep{wu2016parallel} and EI \citep{jones1998efficient,wang2015parallel}, a derivative-enabled single-fidelity version of KG \citep{wu2017bayesian}, Hyperband \citep{li2016hyperband}, and the multi-fidelity method BOCA \citep{kandasamy2017multi}.  BOCA is the only previous method of which we are aware that treats multiple continuous fidelities.  We do not compare against FaBOLAS \citep{klein2016fast,klein2015towards} because the kernel it uses is specialized to neural network hyperparameter tuning.

Following experiments in \citet{kandasamy2017multi}, we augment four synthetic test functions, 2-d Branin, 3-d Rosenbrock, 3-d Hartmann, and 6-d Hartmann, by adding one or two fidelity controls, as described in the supplement.  
We set the cost of an individual evaluation of $x$ at fidelity $s$ to $0.01 + \prod_i s_i$.
Thinking of $s_1$ as mimicking training data and $s_2$ training iterations, 
the term $\prod_i s_i$ mimics a cost of training that is proportional to the number of times a datapoint is visited in training.  The term $0.01$ mimics a fixed cost associated with validating a trained model.
We set the cost of a batch evaluation to the maximum of the costs of the individual evaluations,
to mimic wall-clock time for running evaluations synchronously in parallel. 

\newcommand{\figwidth}{230pt}
\newcommand{\figheight}{190pt}

\begin{figure*}[tb]
\centering
  \subfigure
  \centering
  \includegraphics[width=\figwidth, height = \figheight]{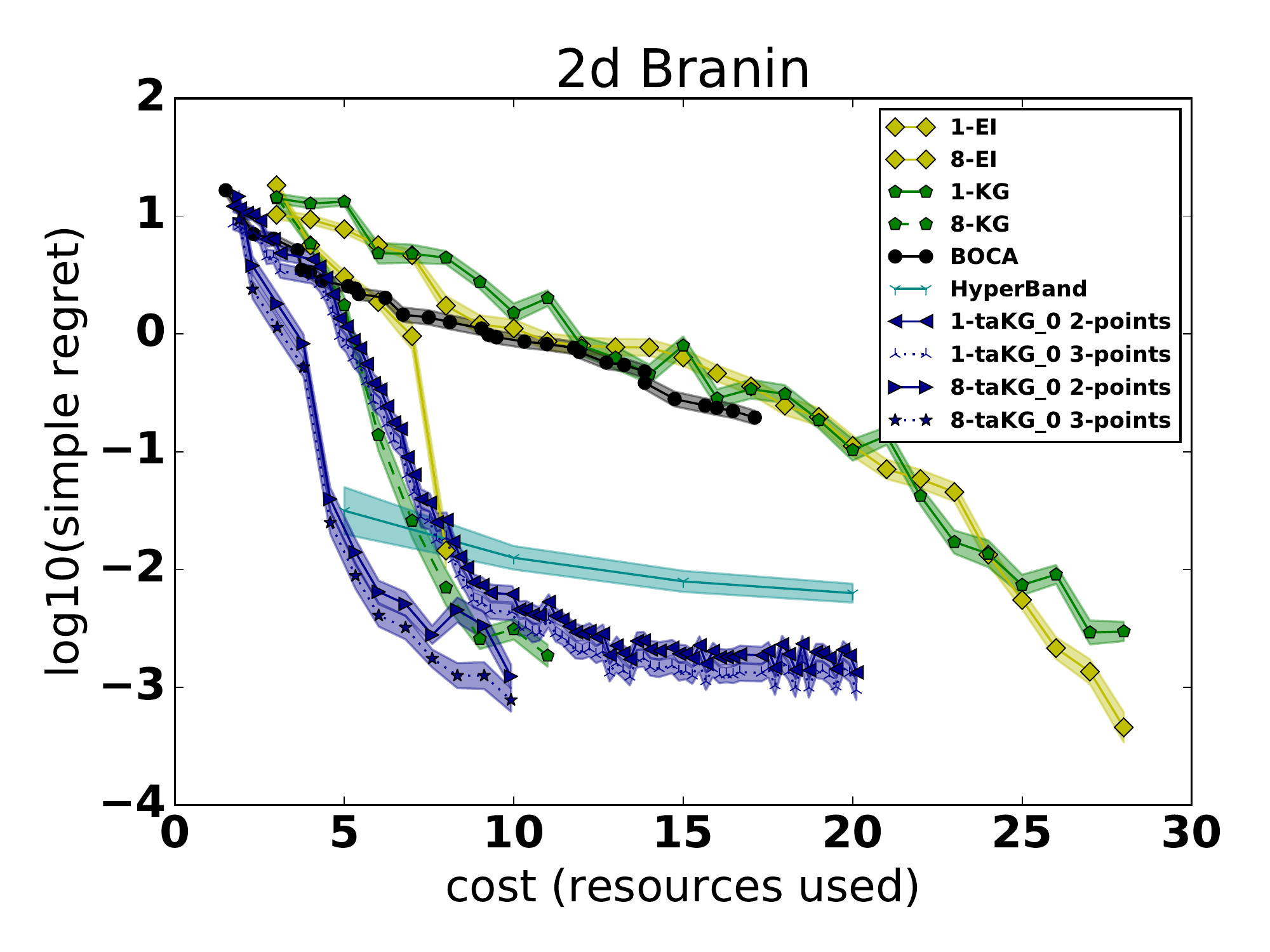}
  \subfigure
  \centering
  \includegraphics[width=\figwidth, height = \figheight]{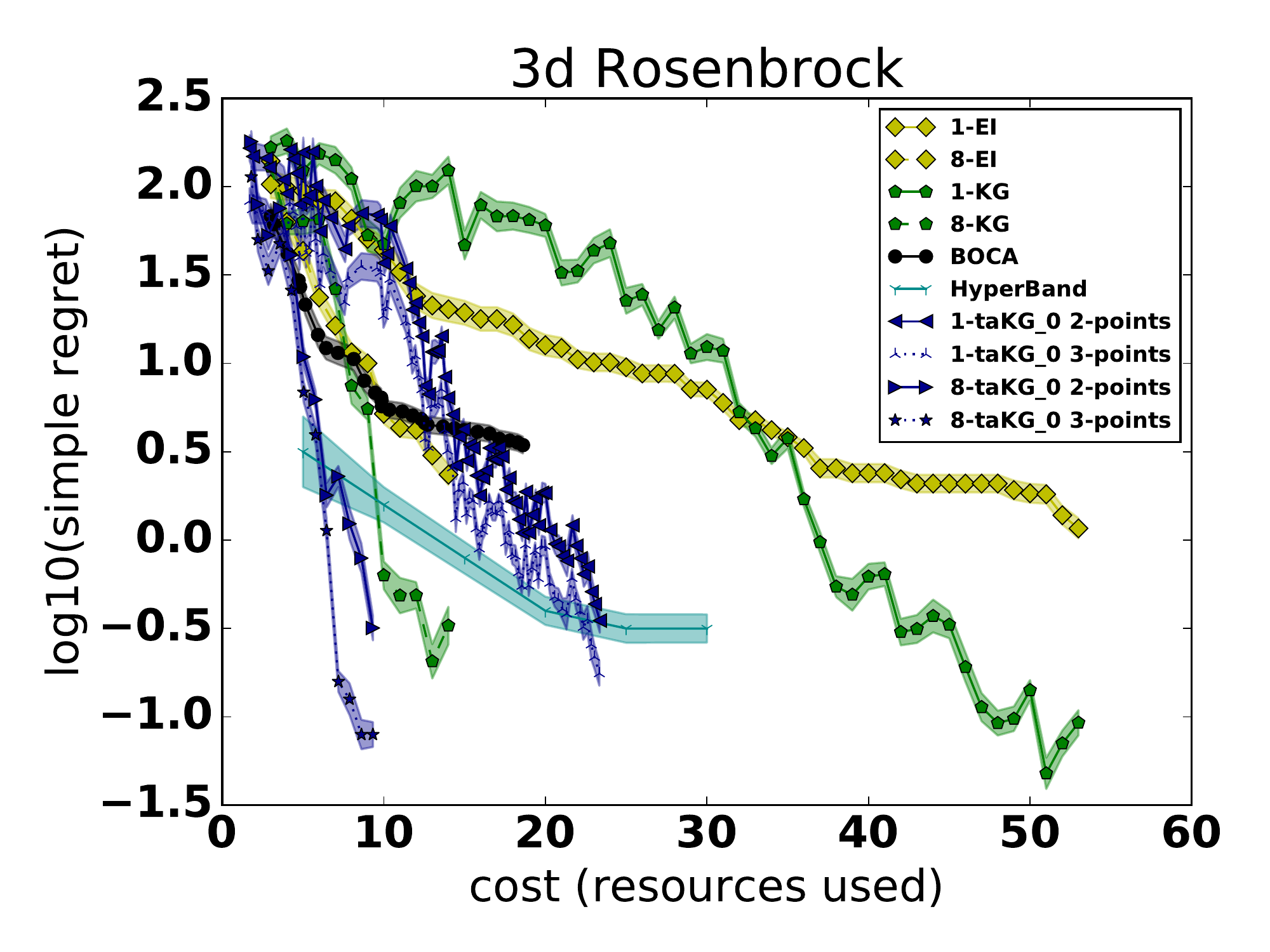}\\
    \subfigure
  \centering
\includegraphics[width=\figwidth, height = \figheight]{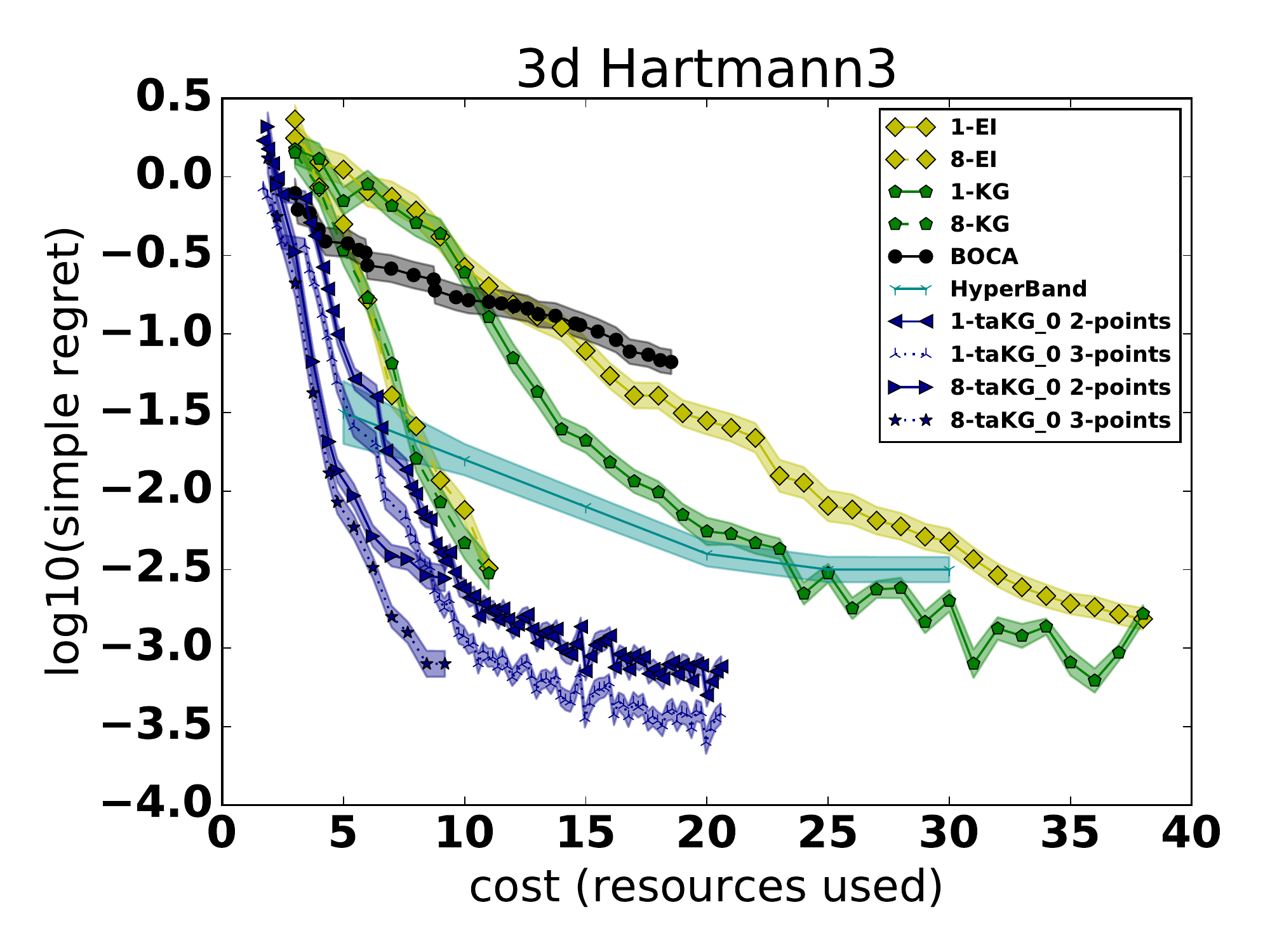}
   \subfigure
  \centering
  \includegraphics[width=\figwidth, height = \figheight]{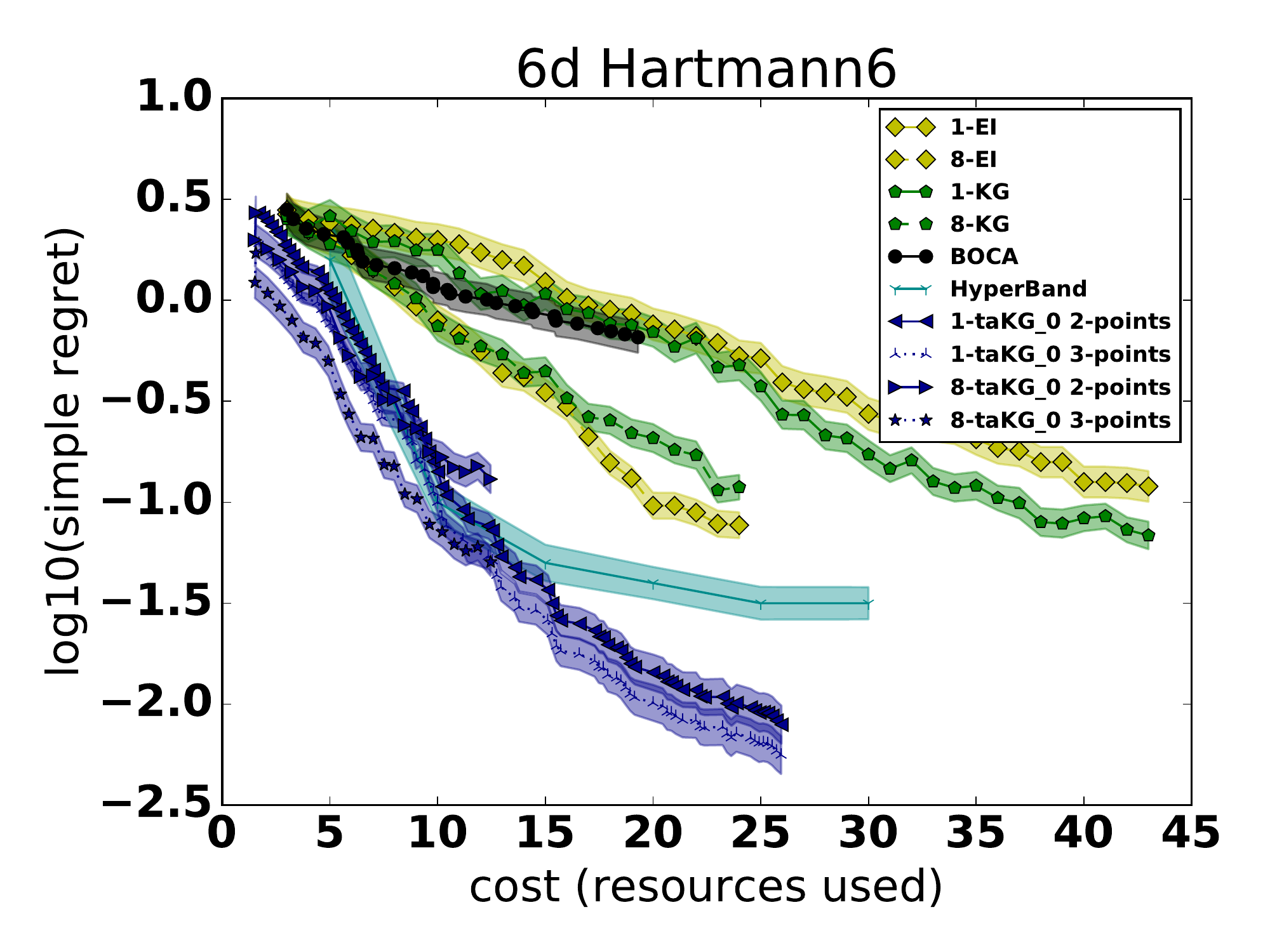}
\vspace{-8pt}
\caption{\small Optimizing synthetic functions: Plots show simple regret over $40$ independent runs for synthetic functions with trace observations and one or two continuous fidelity controls for 2-d Branin, 3-d Rosenbrock, 3-d Hartmann, and 6-d Hartmann problems.  $\taKG0$ performs well compared with a variety of competitors in both sequential and batch settings.}
\vspace{-12pt}
\label{Fig_syn_taKG}
\end{figure*}

Fig.~\ref{Fig_syn_taKG} shows results.  For methods that have both sequential and batch versions, the batch size is indicated with a number before the method name.  For example, 8-EI indicates expected improvement performed with batches of size 8.  We run versions of $\taKGE$ using $|\S|$ set to $2$ (taKG\_0 2-points) and $3$ (taKG\_0 3-points).

We first see that using the larger $|\S|$ improves the performance of $\taKGE$.
We then see that, for both values of $|\S|$, sequential $\taKGE$ performs well relative to sequential competitors (1-EI, 1-KG, 1-BOCA), and batch $\taKGE$ with batch size 8 (8-$\taKGE$) performs well relative to its batch competitors (8-EI, 8-KG, Hyperband).

Here, we consider Hyperband to be a batch method although the amount of parallelism it can leverage varies through the course of its operation.

\subsection{Optimizing hyperparameters of neural networks } \label{sect:hyperexp}
Here, we evaluate on hyperparameter optimization of neural networks. Benchmarks include the single-fidelity Bayesian optimization algorithms KG \citep{wu2016parallel} and EI \citep{jones1998efficient}, their batch versions, and the state-of-art hyperparameter tuning algorithms HyperBand and FaBOLAS.  $\taKGE$ uses two fidelity controls: the size of the training set and the number of training iterations.  

Following \citet{li2016hyperband}, we set the cost to the number of training examples passed during training divided by the number passed in full fidelity. For example, if we have $5\times 10^4$ training points and the maximum number of epochs is $100$, then the cost of evaluating a set of hyperparameters using $10^4$ sub-sampled training points per epoch over $10$ epochs is $10^4 \times 10 / (5\times10^4 \times 100) = 0.02$. Complete training has cost $1$.

\begin{figure*}[tb]
  \centering
  \subfigure
  \centering
  \includegraphics[width=\figwidth, height = \figheight]{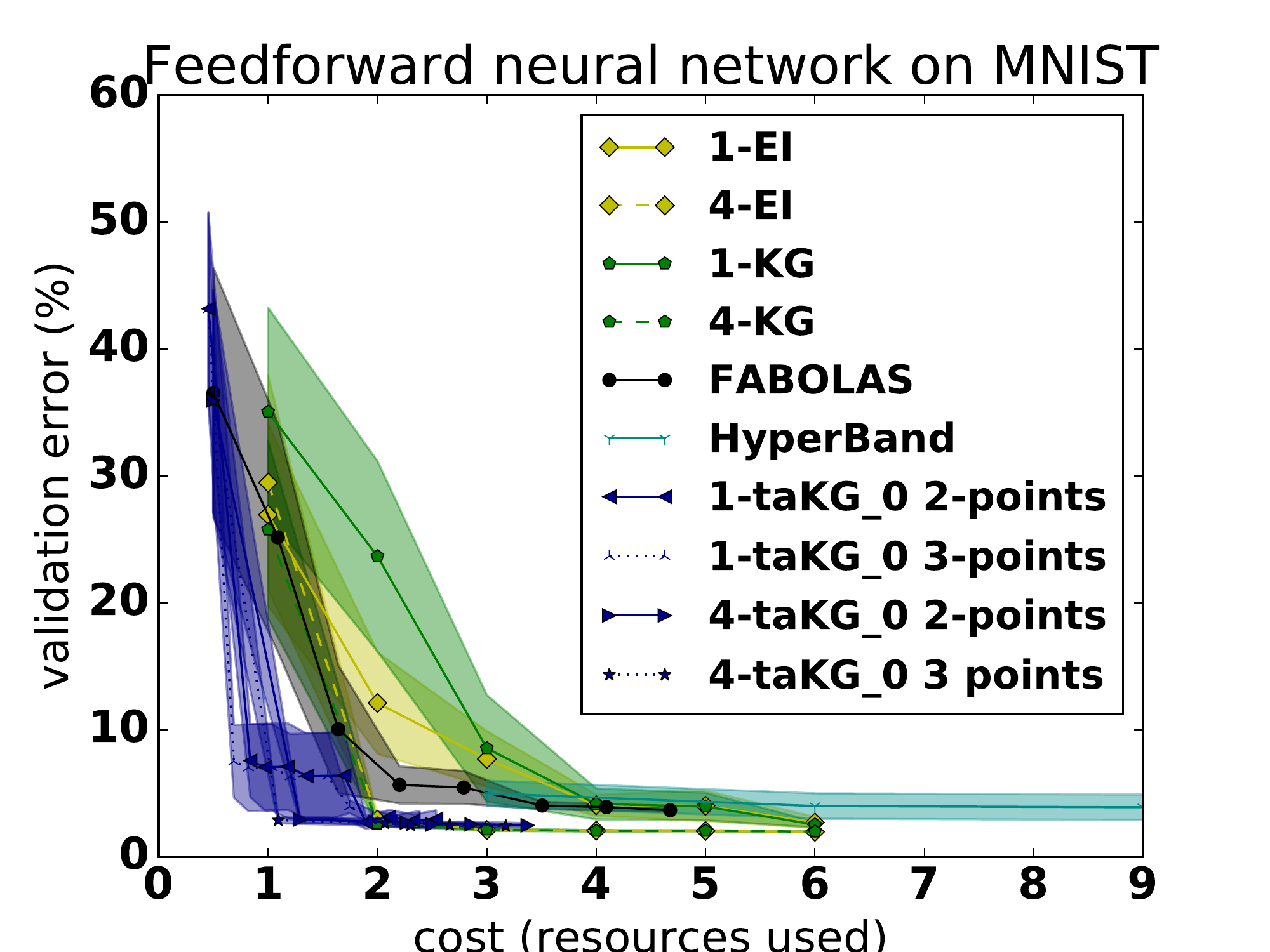}
  \subfigure
  \centering
  \includegraphics[width=\figwidth, height = \figheight]{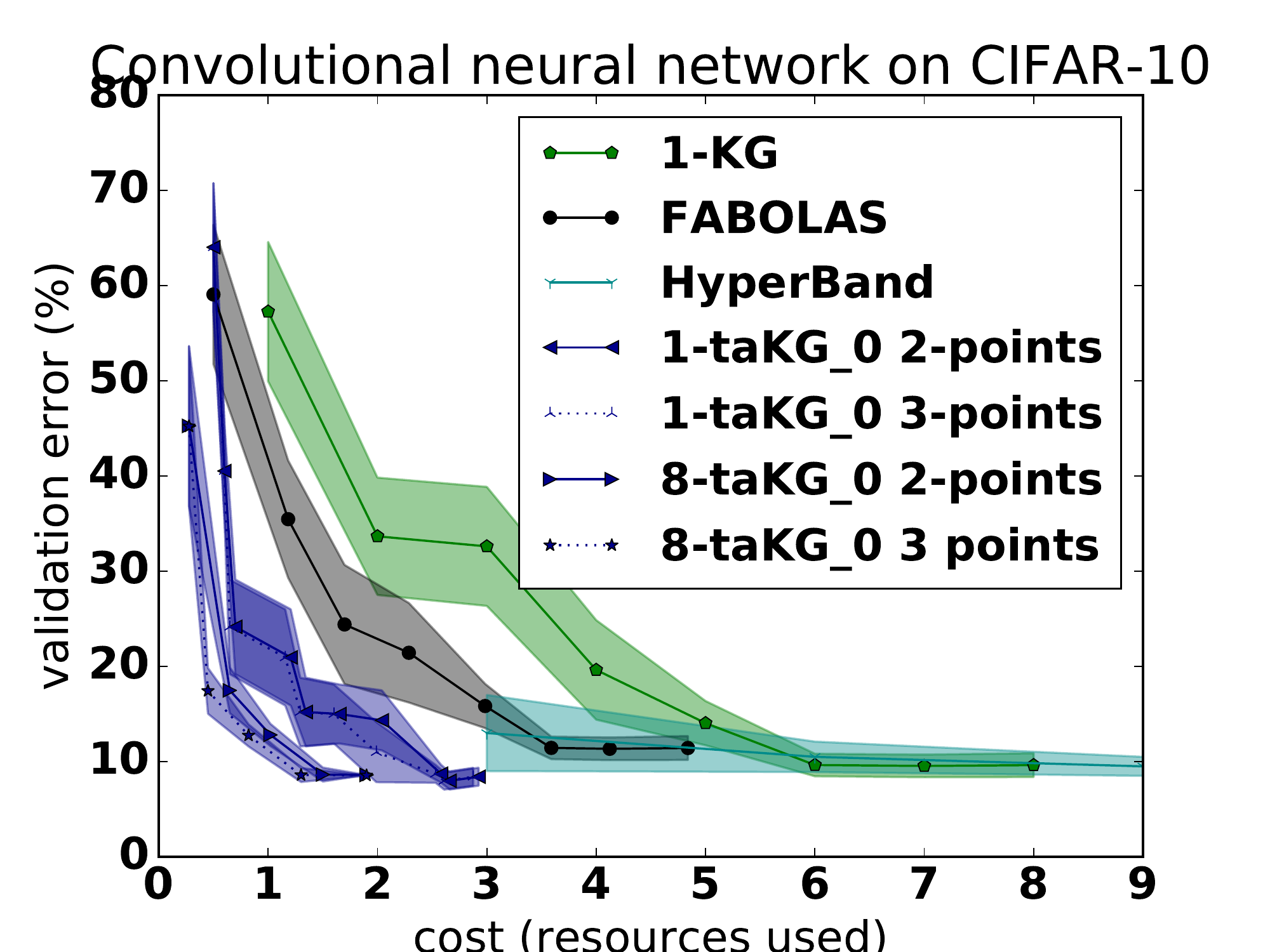}\\
  \subfigure
  \centering
  \includegraphics[width=\figwidth, height = \figheight]{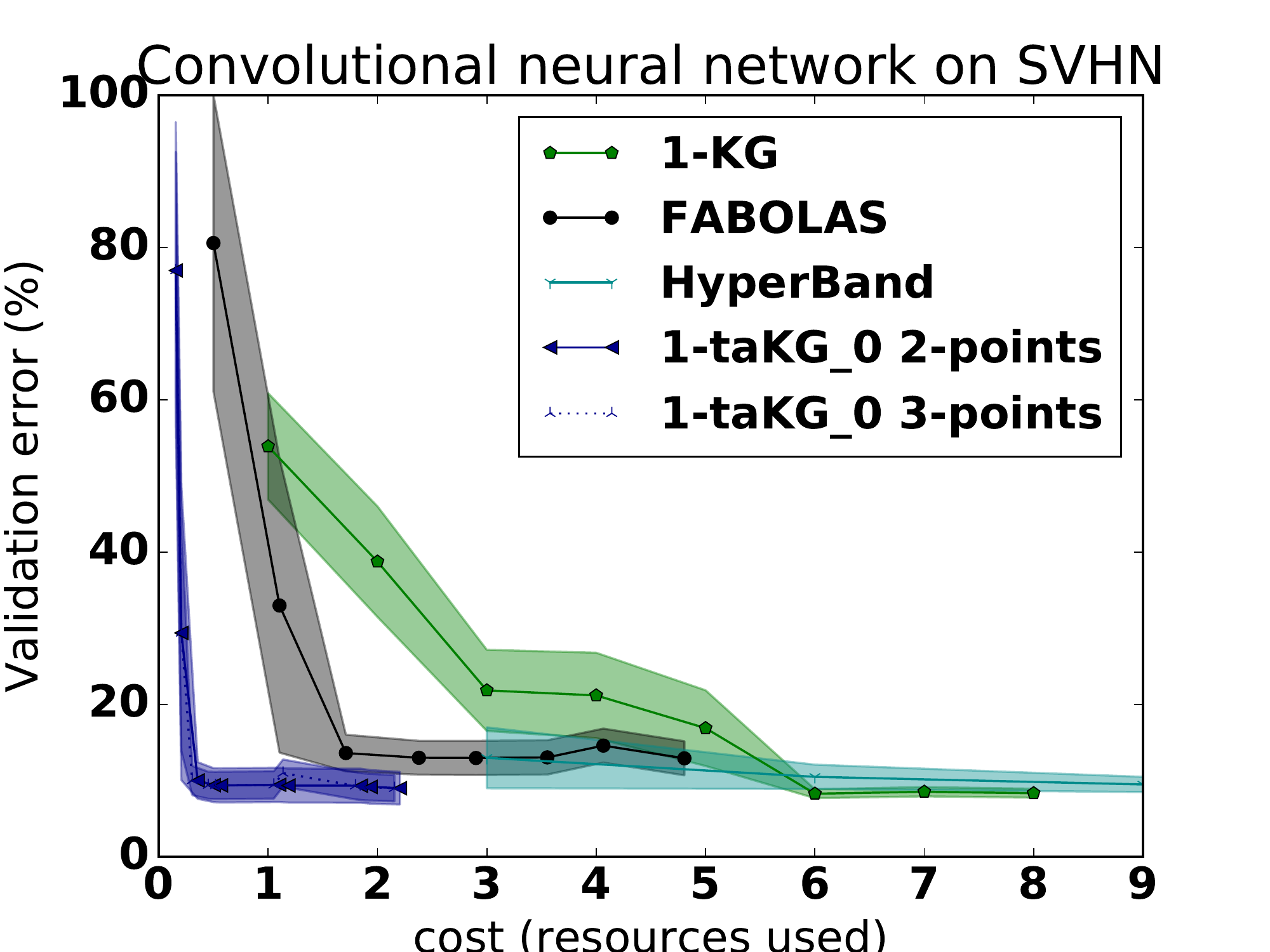}
  \subfigure
  \centering
  \includegraphics[width=\figwidth, height = \figheight]{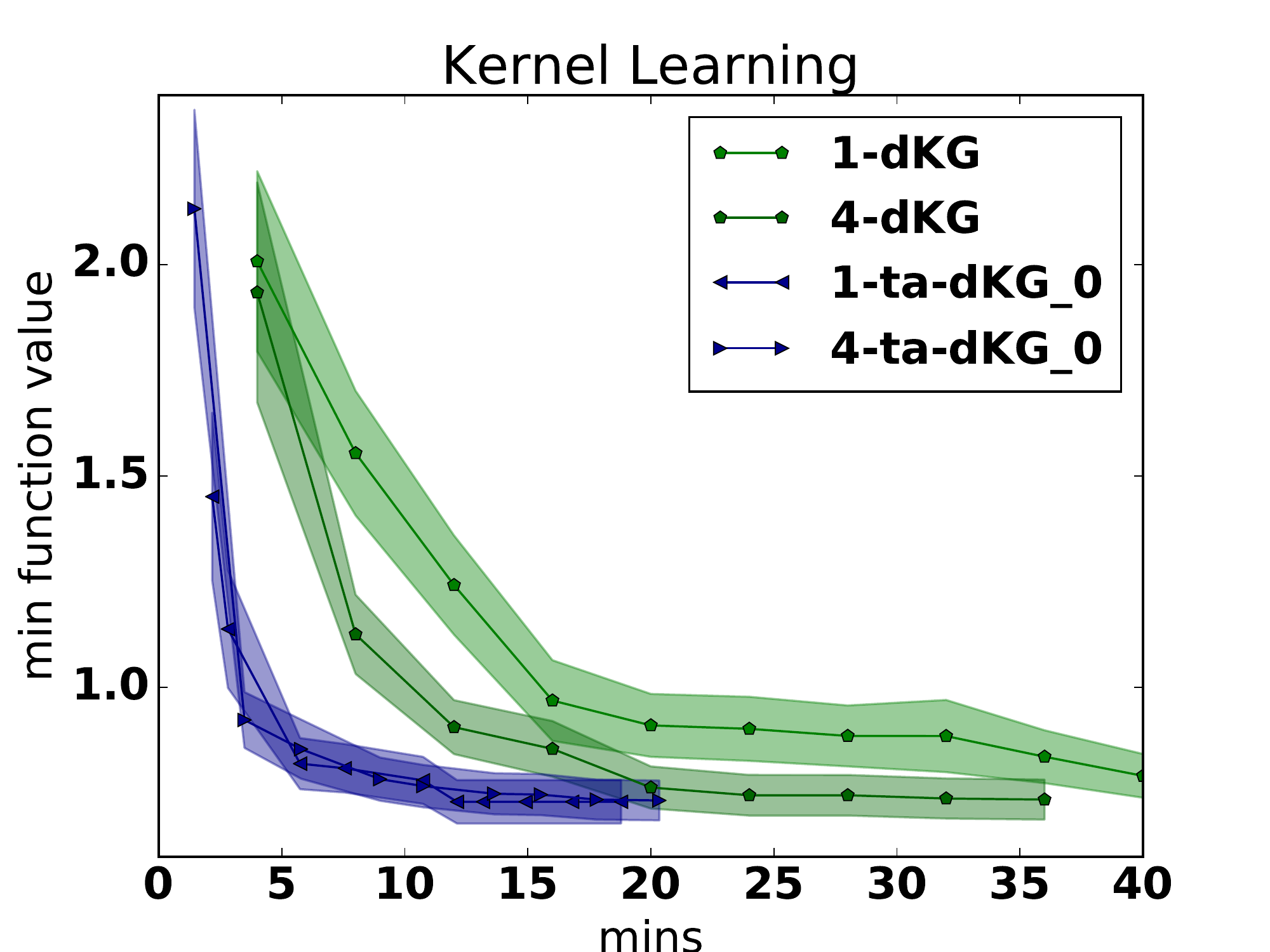}
\vspace{-8pt}
\caption{\small We show the -log marginal likelihood divided by the number of datapoints for tuning feedforward neural networks on MNIST (each with 20 runs); tuning convolutional neural networks on CIFAR-10 and SVHN (each with 10 runs); and for KISS-GP kernel learning.
$\taKGE$ outperforms competitors in both sequential and batch settings.  }
\vspace{-12pt}
\label{Fig_deep_taKG}
\end{figure*}

\paragraph{Feedforward neural networks on MNIST}
\label{sect:mlps}
We tune a fully connected two-layer neural network on MNIST. The maximum number of epochs allowed is $20$. We optimize $5$ hyperparameters: learning rate, dropout rate, batch size and the number of units at each layer. 

Fig.~\ref{Fig_deep_taKG} shows that sequential $\taKGE$ performs much better than the sequential methods KG, EI and the multi-fidelity hyperparameter optimization algorithm FaBOLAS. $\taKGE$ with a batch size $4$ substantially improves over batch versions of KG and EI, and also over the batch method Hyperband.

\paragraph{Convolutional neural networks on CIFAR-10 and SVHN}
\label{sect:cnns}
We tune convolution neural networks (CNNs) on CIFAR-10 and SVHN. Our CNN consists of 3 convolutional blocks and a softmax classification layer. Each convolutional block consists of two convolutional layers with the same number of filters followed by a max-pooling layer. There is no dropout or batch-normalization layer. We split the CIFAR-10 dataset into 40000 training samples, 10000 validation samples and 10000 test samples. We split the SVHN training dataset into 67235 training samples and 6000 validation samples, and use the standard 26032 test samples. We apply standard data augmentation: horizontal and vertical shifts, and horizontal flips. We optimize $5$ hyperparameters to minimize the classification error on the validation set: the learning rate, batch size, and number of filters in each convolutional block. Hyperband uses the size of the training set as its resource (it can use only one resource or fidelity), using a bracket size of $s_{\max} = 4$ as in \citet{li2016hyperband} and the maximum resource allowed by a single configuration set to 40000.
We set the maximum number of training epochs for all algorithms to $50$ for CIFAR-10 and $40$ for SVHN. 
Because of the computational expense of training CNNs, we leave out some benchmarks, dropping the single-fidelity method EI in favor of the structurally similar single-fidelity method KG, and performing batch evaluations for only some methods.

Fig.~\ref{Fig_deep_taKG} shows that sequential $\taKGE$ outperforms its competitors (including the batch method Hyperband) on both problems.  Using batch evaluations with $\taKGE$ on CIFAR-10 improves performance even further.

When we train using optimized hyperparameters on the full training dataset for $200$ epochs, test data classification error is $\sim12\%$ for CIFAR-10 and $\sim5\%$ for SVHN.

\subsection{Optimizing hyperparameters for large-scale kernel learning}
\label{sect:kernel-learning}
We test derivative-enabled $\taKGE$ (ta-dKG$^\emptyset$) in a large-scale kernel learning example: the 1-d demo example for KISS-GP \citep{wilson2015kernel} on the GPML website \citep{gpml}.  In this example, we optimize 3 hyperparameters (marginal variance, length scale, and variance of the noise) of a GP with an RBF kernel on $1$ million training points to maximize the log marginal likelihood. We evaluate both the log marginal likelihood and its gradient using the KISS-GP framework. We use two continuous fidelity controls: the number of training points and the number of inducing points.  We set the maximum number of inducing points to $m=1000$.

We compare ta-d-KG to the derivative-enabled knowledge gradient ($\qKGg$) \citep{wu2017bayesian}, using both algorithms in the sequential setting (1-dKG and 1-cf-dKG) and with a batch size of 4 (4-dKG and 4-cf-dKG).  We leave out methods that are unable to utilize derivatives, as these are likely to substantially underperform.

Fig.~\ref{Fig_deep_taKG} shows that ta-dKG$^\emptyset$ successfully utilizes inexpensive function and gradient evaluations to find a good solution more quickly than d-KG, in both the sequential and batch setting.



\section{CONCLUSION}
\label{sect:conclusion}
We propose a novel multi-fidelity acquisition function, the trace aware knowledge-gradient, which 
leverages special structure provided by trace observations,
is able to handle multiple simultaneous continuous fidelities, and 
generalizes naturally to batch and derivative settings.
This acquisition function uses traces to find 
good solutions to global optimization problems more quickly than state-of-the-art algorithms
in application settings including deep learning and kernel learning.



\bibliographystyle{apalike}
\bibliography{pKG}

\newpage


\section{SUPPLEMENTARY MATERIAL}

\subsection{Background: Gaussian processes}
\label{sect:gp}

We put a Gaussian process (GP) prior \citep{rasmussen2006gaussian} on the function $g$.  
The GP prior is defined by its mean function $\mu_{0}: \mathbb{A} \times [0, 1]^m \mapsto \mathbb{R}$ and kernel function $K_{0}: \left\{\mathbb{A} \times [0, 1]^m\right\} \times \left\{\mathbb{A} \times [0, 1]^m \right\} \mapsto \mathbb{R}$.  These mean and kernel functions have hyperparameters, whose inference we discuss below.

We assume that evaluations of $g(x,s)$ are subject to additive independent normally distributed noise with common variance $\sigma^2$.
We treat the parameter $\sigma^2$ as a hyperparameter of our model, and also discuss its inference below.  Our assumption of normally distributed noise with constant variance is common in the BO literature \citep{klein2016fast}. 

The posterior distribution on $g$ after observing $n$ function values at points $z_{(1:n)} := \{(x_{(1)}, s_{(1)}), (x_{(2)}, s_{(2)}), \cdots, (x_{(n)}, s_{(n)})\}$ with observed values $y_{(1:n)} := \{y_{(1)}, y_{(2)}, \cdots, y_{(n)}\}$ remains a Gaussian process \citep{rasmussen2006gaussian}, and $g \mid z_{(1:n)}, y_{(1:n)} \sim \text{GP}(\mu_n, K_n)$ with $\mu_n$ and $K_n$ evaluated at a point $z=(x,s)$ (or pair of points $z$, $\tilde{z} = (\tilde{x},\tilde{s})$) given as follows

\begin{equation}
\begin{split}
& \bm\mu_n(\bm{x}) = \mu(\bm{x})\\
&\qquad + K(\bm{x},x_{1:n}) \left(K(x_{1:n},x_{1:n})+\right.\\
&\qquad \left.\sigma^2 I \right)^{-1}  (y_{1:n}-\mu(x_{1:n})),\\
& K_{n}(\bm{x_1}, \bm{x_2}) =  K(\bm{x_1},\bm{x_2})\\
&\qquad - K(\bm{x_1},x_{1:n})  \left(K(x_{1:n},x_{1:n})\right.\\
&\qquad\left.+\sigma^2 I\right)^{-1} K(x_{1:n},\bm{x_2}).
\end{split}
\label{eq:posterior}
\end{equation}

We emphasize that a single evaluation of $g$ provides multiple observaions of $g$, because of trace observations.  taKG chooses to retain 2 observations per evaluation, and so $n$ will be twice the number of evaluations.


This statistical approach contains several hyperparameters: the variance $\sigma^2$, and any parameters in the mean and kernel functions.  We treat these hyperparameters in a Bayesian way as proposed in \citet{snoek2012practical}. We analogously train a separate GP on the logarithm of the cost of evaluating $g(x,s)$.

\subsection{Proofs Details}

In this section we prove the theorems of the paper. We may assume without loss of generality that $|\mathcal{S}| = 1$, and we denote the number of fidelities by $m$. Observe that the dimension of the vector  $C(\mathcal{S}) \cup \mathcal{S}$ is $q:=2m$.
We first show some smoothness properties of $\tilde{\sigma}_{n}$, $\mu_{n}$ and  $\mbox{cost}_{n}$ in the following lemma.
\begin{lemma}
\label{smooth_lemma}
We assume that the domain  $\mathbb{A}$ is compact, $\mu_0$ is a constant and the kernel $K_{0}$ is continuously differentiable. We then have that
\begin{enumerate}
\item $\mu_{n}$, and $\tilde{\sigma}_{n}\left(\cdot,z_{1:q}\right)$ are both continuously differentiable for any vector $z_{1:q}$.
\item For any $x'$, $\tilde{\sigma}_{n}\left(x',z_{1:q}\right)$ is continuously  differentiable respect to $z_{1:q}$.
\item $\mbox{cost}_{n}$ is continuously differentiable.
\item $\max_{1 \le i \le q}\text{cost}_{n}(x_i, s_i)$ is differentiable if $\left|\mbox{arg max}_{1\leq i\leq q}\mbox{cost}_{n}\left(x_{i},s_{i}\right)\right|=1$.
\end{enumerate}
\end{lemma}
\proof{}
The posterior parameters of the Gaussian process $\mu_{n}$, $K_{n}$ and $\mbox{cost}_{n}$ are continuously differentiable if the kernel $K_{0}$ and $\mu_{0}$ are both continuously differentiable. Observe that $\tilde{\sigma}_n(x, z_{(1:q)})= K_{n}\left((x, 1), z_{1:q}\right) (D_{n}\left(z_{1:q}\right)^T)^{-1}$, and so $\tilde{\sigma}_{n}\left(\cdot,z_{1:q}\right)$ is continuously differentiable because $K_{n}$ is continuously differentiable. This proves (1) and (3). (4) follows easily from (3).

To prove (2) we only need to show that $(D_{n}\left(z_{1:q}\right)^T)^{-1}$ is continuously differentiable respect to $z_{1:q}$. This follows from the fact that multiplication, matrix inversion (when the inverse exists), and Cholesky factorization \citep{smith1995differentiation} preserve continuous differentiability. This ends the proof.
\endproof

We now prove Theorem 1.
\proof{}
Let $W_q$ be a standard normal random vector, and define $f(x,z_{1:q}):=\mu_n \left(x, 1\right)+\tilde{\sigma}_n\left(x, z_{1:q}\right)W_q$. By Lemma \ref{smooth_lemma}, $f$ is continuously differentiable. By the envelope theorem (see Corollary 4 of \citealt{milgrom2002envelope}), $\nabla f(y,z_{1:q})=\nabla \tilde{\sigma}_n\left(y, z_{1:q}\right)W_q$ a.s., where $y=\mbox{arg max}_{x\in \mathbb{A}}\left(\mu_n(x, 1_q)+\tilde{\sigma}_n\left(x, z_{1:q}\right)W_{q}\right)$.

We now show that we can interchange the gradient and the expectation in $\nabla \mathbb{E}_n\left[\min_{x \in \mathbb{A}} \left(\mu_{n} \left(x, 1\right)+\tilde{\sigma}_n\left(x, z_{1:q}\right)W_q\right)\right]$. Observe that the domain of $z_{1:q}$ is compact,  $\tilde{\sigma}_{n}\left(x',z_{1:q}\right)$ is continuously  differentiable respect to $z_{1:q}$ by Lemma \ref{smooth_lemma}, and so $\left\Vert\tilde{\sigma}_{n}\left(x',z_{1:q}\right)\right\Vert$ is bounded. Consequently, Corollary 5.9 of \citet{bartle} implies that we can interchange the gradient and the expectation. The statement of the theorem follows from the strong law of large numbers.
\endproof

The following corollary follows from the previous proof.

\begin{corollary}
Under the assumptions of the previous theorem, $L_{n}(x,S)$ is continuous.
\end{corollary}

We now prove Theorem 2.
\proof{}
We prove this theorem using Theorem 2.3 of Section 5 of \citet{kushner2003stochastic}, which depends on the structure of the stochastic gradient  $G$ of the objective function. In addition, we simplify the notation and denote $(X,\S)$  by $Z$.



The theorem from \citet{kushner2003stochastic}, requires the following hypotheses:
\begin{enumerate}
\item $\epsilon_{t}\rightarrow0$, $\sum_{t=1}^{\infty}\epsilon_{t}=\infty$,
and $\sum_{t}\epsilon_{t}^{2}<\infty$.
\item $\sup_{t}E\left[\left|G\left(Z_{t}\right)\right|^{2}\right]<\infty$
\item There exist uniformly continuous functions $\left\{ \lambda_{t}\right\} _{t\geq0}$
of $Z$, and random vectors $\left\{ \beta_{t}\right\} _{t\geq0}$
, such that $\beta_{t}\rightarrow0$ almost surely and
\[
E_{n}\left[G\left(Z_{t}\right)\right]=\lambda_{t}\left(Z_{t}\right)+\beta_{t}.
\]
Furthermore, there exists a continuous function $\bar{\lambda}$,
such that for each $Z\in A^{q}$,
\[
\lim_{n}\left|\sum_{i=1}^{m\left(r_{m}+s\right)}\epsilon_{i}\left[\lambda_{i}\left(Z\right)-\bar{\lambda}\left(Z\right)\right]\right|=0
\]
for each $s\geq0$, where $m\left(r\right)$ is the unique value of
$k$ such that $t_{k}\leq t<t_{k+1}$, where $t_{0}=0$,$t_{k}=\sum_{i=0}^{k-1}\epsilon_{i}$.
\item There exists a continuously differentiable real-valued function $\phi$,
such that $\bar{\lambda}=-\nabla\phi$ and it is constant on each
connected subset of stationary points.
\item The constraint functions defining $\mathbb{A}$ are continuously differentiable.
\end{enumerate}
We now prove that our problem satisfy these hypotheses. (1) is true
by hypothesis of the lemma. 

Let's prove (2). We first assume that $r=1$, 
\begin{eqnarray*}
E\left[\left|\frac{1}{r}\sum_{k=1}^{r}\nabla\tilde{\sigma}_{n}\left(y_{k},Z\right)W_{q}^{k}\right|^{2}\right] & =\\
E\left[\left|\nabla\tilde{\sigma}_{n}\left(y_{1},Z\right)W_{q}^{1}\right|^{2}\right] & \leq\\
E\left[\left\Vert \nabla\tilde{\sigma}_{n}\left(y_{1},Z\right)\right\Vert ^{2}\left\Vert W_{q}^{1}\right\Vert ^{2}\right] & \leq\\
qL
\end{eqnarray*}
where $L:=\mbox{sup}_{x,z}\left\Vert \nabla\tilde{\sigma}_{n}\left(x,z\right)\right\Vert ^{2}$,
which is finite because the domain of the problem is compact and $\nabla\tilde{\sigma}_{n}\left(x,z\right)$
is continuous by Lemma 1. Since $C:=\mbox{cost}^{\left(n\right)}$ is continuously
differentiable bounded below by a constant $K$, thus we conclude
that the supremum over $Z$ of $E\left[\left|G\left(Z\right)\right|^{2}\right]$
is bounded. If $r>1$, $G\left(Z_{n}\right)$ is the average of i.i.d.
random vectors, whose squared norm expectation is finite, and so $\sup_{t}E\left[\left|G\left(Z_{t}\right)\right|^{2}\right]$
must be finite too.

We now prove (3). For each $t$, define 
\begin{eqnarray*}
\lambda_{t}\left(Z\right) & =\\
E\left[\frac{C\left(Z\right)\nabla\tilde{\sigma}_{n}\left(Y,Z\right)W_{q}}{C\left(Z\right)^{2}}\right]\\
-E\left[\frac{\nabla C\left(Z\right)}{C\left(Z\right)^{2}}\left(\mu_{n}\left(Y,1_{m}\right)+\tilde{\sigma}_{n}\left(Y,Z\right)W_{q}\right)\right]
\end{eqnarray*}
 where $Y=\mbox{arg max}_{x}\left(\mu_{n}\left(x\right)+\tilde{\sigma}_{n}\left(x,Z\right)W_{q}\right)$,
and $W_{q}$ is a standard normal random vector. Let's prove that
$\lambda_{t}$ is continuous. In the proof of Theorem 1, we show that
$\nabla\tilde{\sigma}_{n}\left(Y,Z\right)W_{q}$ is continuous in
$Z$. Furthermore,
\begin{eqnarray*}
\left\Vert \nabla\tilde{\sigma}_{n}\left(y_{1},Z\right)W_{q}^{1}\right\Vert  & \leq & \left\Vert \nabla\tilde{\sigma}_{n}\left(y_{1},Z\right)\right\Vert \left\Vert W_{q}\right\Vert \\
 & \leq & L\left\Vert W_{q}\right\Vert .
\end{eqnarray*}
Consequently $E\left[\nabla\tilde{\sigma}_{n}\left(Y,Z\right)W_{q}\right]$
is continuous by Corollary 5.7 of \citet{bartle}. In Theorem 1, we also show
that $E\left[\left(\mu_{n}\left(Y,1_{m}\right)+\tilde{\sigma}_{n}\left(Y,Z\right)W_{q}\right)\right]$
is continuous in $Z$. Since $C$ is continuously differentiable,
we conclude that $\lambda_{t}$ is continuous. By defining $\beta_{t}=0$
for all $t$, and $\bar{\lambda}=\lambda_{1}$, we conclude the proof
of (3).

Finally, define $\phi\left(Z\right)=-E\left[\frac{\mu_{n}\left(Y,1_{m}\right)+\tilde{\sigma}_{n}\left(Y,Z\right)W_{q}}{C\left(Z\right)}\right]$.
Observe that in Lemma 2, we show that we can interchange the expectation
and the gradient in $E\left[\nabla\left(\mu_{n}\left(Y\right)+\tilde{\sigma}_{n}\left(Y,Z\right)W_{q}\right)\right]$,
and so $\lambda_{m}\left(Z\right)=-\nabla\phi\left(Z\right).$ In
a connected subset of stationary points, we have that $\lambda_{m}\left(Z\right)=0$,
and so $\phi\left(Z\right)$ is constant. This ends the proof of the
theorem.
\endproof

\begin{proof}[proof of Proposition 1]
Since
\begin{eqnarray*}
VOI_{n}(x, s) &:=& \\
\E_n[\mu^*(x, 1) - \min_{x'} (\mu_n(x') + C_{n}(x', (x, s)) W]
\end{eqnarray*}
where $W$ is a standard normal random variable. By Jensen's inequality, we have
\begin{eqnarray*}
VOI_{n}(x, s) &:=& \E_n[\mu^*(x, 1) - \min_{x'} (u_{n}\left(x',s,W\right))]\\
&\ge& \mu^*(x, 1) - \min_{x'} \E_n(u_{n}\left(x',s,W\right)) = 0.
\end{eqnarray*}
where $u_{n}\left(x,s,W\right):=\mu^n(x', 1) + C_{n}((x', 1), (x, s)) W)$.
The inequality becomes equal only if $\min_{x'} (\mu_n(x') + C_{n}(x', (x, s)) W$ is a linear function of $W$ for any fixed $(x, s)$, i.e the argmin for the inner optimization function doesn't change as we vary $W$, which is not true if $K_{n}((x', 1), (x, s))>0$ i.e. evaluating at $(x, s)$ provides value to determine the argmin of the surface $(x, 1)$.
\end{proof}

\begin{proof}[proof of Proposition 2]
The proof follows a very similar argument than the previous proof. By Jensen's inequality, we have that

\begin{eqnarray*}
\E_{n}\left[\min_{x'}\E_{n}\left[g(x',1)\mid\Y(x,\S)\right]\right] & \geq\\
\E_{n}\left[\min_{x'}\E_{n}\left[g(x',1)\mid\Y(x,\S\bigcup C(\S))\right]\right]
\end{eqnarray*}

The inequality becomes equal only if the argmin for the inner optimization function doesn't change as we vary the normal random vector, which is not true under our assumptions.

\end{proof}

\subsection{GPs for Hyperparameter Optimization}
\label{sect:gp-ho}
In the context of hyperparameter optimization with two continuous fidelities, i.e. the number of training iterations ($s_{(1)}$) and the amount of training data ($s_{(2)}$), we set the kernel function of the GP as
\begin{eqnarray*}
K_{0}(z, \tilde z) = K(x, \tilde x) \times K_1(s_{(1)}, \tilde s_{(1)}) \times K_2(s_{(2)},, \tilde s_{(2)}),
\end{eqnarray*}
where $K(\cdot, \cdot)$ is a square-exponential kernel. If we assume that the learning curve looks like
\begin{eqnarray}
g(x, s) = h(x) \times \left(\beta_0 + \beta_1\exp{(-\lambda s_{(1)})}\right) \times l(s_{(2)}),
\label{eqn:shape}
\end{eqnarray}
then inspired by \cite{swersky2014freeze}, we set the kernel $K_1(\cdot, \cdot)$ as
\begin{eqnarray*}
K_1(s_{(1)}, \tilde s_{(1)}) = \left( w + \frac{\beta^{\alpha}}{(s_{(1)} + \tilde s_{(1)} + \beta^{\alpha})}\right),
\end{eqnarray*}
where $w, \beta, \alpha > 0$ are hyperparameters. We add an intercept $w$ compared to the kernel in \cite{swersky2014freeze} to model the fact that the loss will not diminish. We assume that the kernel $K_2(\cdot, \cdot)$ has the form
\begin{eqnarray*}
K_2(s_{(2)}, \tilde s_{(2)}) = \left(c + (1-s_{(2)})^{(1+\delta)}(1-\tilde s_{(2)})^{(1+\delta)}\right),
\end{eqnarray*}
where $c, \delta > 0$ are hyperparameters.

All the hyperparameters can be treated in a Bayesian way as proposed in \citet{snoek2012practical}.

\subsection{Additional experimental details}
\label{sect:addition}
\subsubsection{Synthetic experiments}
Here we define in detail the synthetic test functions on which we perform numerical experiments
The test functions are: 
\begin{eqnarray*}
&&\text{augmented-Branin}(x, s) \\
&=& \left(x_2 - \left(\frac{5.1}{4 \pi^2} - 0.1*(1-s_1)\right) x_1^2 + \frac{5}{\pi} x_1 - 6\right)^2 \\
&& \quad + 10*\left(1-\frac{1}{8\pi}\right)\cos(x_1) + 10\\
&&\text{augmented-Hartmann}(x, s) \\
&=& \left(\alpha_1-0.1*(1-s_1)\right) \exp{\left(-\sum_{j=1}^d A_{ij} (x_j - P_{1j})^2\right)}\\
&&\quad + \sum_{i=2}^{4} \alpha_i \exp{\left(-\sum_{j=1}^d A_{ij} (x_j - P_{ij})^2\right)}\\
&&\text{augmented-Rosenbrock}(x, s) \\
& = & \sum_{i=1}^2 \left(100*(x_{i+1}-x_i^2 + 0.1*(1-s_1))^2\right.\\
& & + \left.\left(x_i-1+0.1*(1-s_2)^2\right)^2\right).
\end{eqnarray*} 

\subsubsection{Real-world experiments}
The range of search domain for feedforward NN experiments: the learning rate in $[10^{-6}, 10^0]$, dropout rate in $[0, 1]$, batch size in $[2^5, 2^{10}]$ and the number of units at each layer in $[100, 1000]$.

The range of search domain for CNN experiments: the learning rate in $[10^{-6}, 1.0]$, batch size $[2^5, 2^{10}]$, and number of filters in each convolutional block in $[2^5, 2^9]$.

\bibliography{supplement}

\end{document}